# A Secured Triad of IoT, Machine Learning, and Blockchain for Crop Forecasting in Agriculture[*]

Najmus Sakib Sizan[1], Md. Abu Layek[1†], and Khondokar Fida Hasan[3‡]

[1] Department of Computer Science and Engineering, Jagannath University, Dhaka, Bangladesh
nsakibsizan115@gmail.com
[2] Department of Computer Science and Engineering, Jagannath University, Dhaka, Bangladesh
layek@cse.jnu.ac.bd
[3] Cybersecurity Discipline, School of Professional Studies, University of New South Wales (UNSW), ACT, Australia fida.hasan@unsw.edu.au

**Abstract.** To improve crop forecasting and provide farmers with actionable data-driven insights, we propose a novel approach integrating IoT, machine learning, and blockchain technologies. Using IoT, real-time data from sensor networks continuously monitor environmental conditions and soil nutrient levels, significantly improving our understanding of crop growth dynamics. Our study demonstrates the exceptional accuracy of the Random Forest model, achieving a 99.45% accuracy rate in predicting optimal crop types and yields, thereby offering precise crop projections and customized recommendations. To ensure the security and integrity of the sensor data used for these forecasts, we integrate the Ethereum blockchain, which provides a robust and secure platform. This ensures that the forecasted data remain tamper-proof and reliable. Stakeholders can access real-time and historical crop projections through an intuitive online interface, enhancing transparency and facilitating informed decision-making. By presenting multiple predicted crop scenarios, our system enables farmers to optimize production strategies effectively. This integrated approach promises significant advances in precision agriculture, making crop forecasting more accurate, secure, and user-friendly.

**Keywords:** Industry 5.0 · Internet of Things · Machine Learning · Blockchain technology

## 1 Introduction

Agriculture has always been a vital sector that directly affects the global economy and food security. As the world population continues to grow, the demand for agricultural products increases, necessitating advancements in farming practices. Traditional farming methods, which are based on farmers' experience and intuition, are often insufficient to meet the modern demands for efficiency and sustainability. The integration of advanced technologies such as the Internet of Things (IoT), machine learning,

---

and blockchain into agriculture represents a significant shift towards precision farming. These technologies enable real-time monitoring, predictive analytics, and secure data management, collectively driving smarter and more efficient agricultural practices.

At the heart of Industry 5.0 lies a triad of core values: human-centricity, sustainability, and resilience. This paradigm promotes a production ethos that pivots around human needs, champions environmental sustainability through circular practices, and fosters resilience to adeptly navigate crises and vicissitudes [1].

This research makes several significant contributions to the field of precision agriculture. Firstly, it demonstrates the efficacy of the Random Forest model in achieving an impressive 99.45% accuracy rate for crop forecasting. Secondly, it highlights the integration of Ethereum blockchain to secure sensor data, ensuring its integrity and trustworthiness. In addition, the study presents an intuitive online interface for stakeholders to access current crop projections and historical data, facilitating informed decision making and transparency. By combining these technologies, our approach offers a comprehensive solution that addresses both the predictive and security needs of modern agriculture.

The paper is structured as follows. Section 1 introduces the importance of integrating IoT, machine learning, and blockchain in agriculture. Section 2 provides a detailed review of related work and existing technologies in precision farming. Section 3 outlines the methodology, including the design and implementation of the IoT sensor network, the Random Forest model for crop prediction, and the Ethereum blockchain for data security. Section 4 presents the experimental results, emphasizing the accuracy of the model and the overall performance of the system. Finally, Section 5 concludes the paper, summarizing the key contributions and suggesting directions for future research.

## 2    Literature Review

The convergence of agriculture, technology, and data science has led to notable advances in smart farming practices. This review of the literature synthesizes recent research efforts focused on the use of statistical techniques, machine learning algorithms, and blockchain technology to improve diverse aspects of agricultural systems, including crop forecasting, precision agriculture, security monitoring, yield estimation, and supply chain traceability.

Bhuyan et al. (2022) used statistical and machine learning methodologies to predict crop types in a smart Indian city. Their study comprehensively evaluated algorithms such as k-nearest Neighbors (k-NN), Support Vector Machines (SVM), Random Forests (RF), and Gradient Boosting (GB) trees on a dataset comprising 22 crops and environmental factors. Employing Grey forecasting and cross-validation techniques ensured dataset balance and predictive accuracy [2] .

Rahman et al. (2020) examined obstacles to blockchain integration in smart agriculture and proposed a scalable data-sharing system. Their framework included layers for smart agriculture, smart contracts, the InterPlanetary File System (IPFS), and stakeholders, prioritizing user privacy and utilizing the EOS blockchain [4].

Chaganti et al. (2022) introduced a cloud-enabled smart farm security monitoring framework that incorporates behavioral pattern analysis, blockchain-based smart con-tracts, GPS sensors for animal tracking, and pressure sensors for water management. Using the Arduino Sensor Kit, ESP32, AWS cloud, and Ethereum Rinkeby Test Network, their system bolsters agriculture security monitoring while ensuring data integrity [5].

Pradhan et al. (2021) introduced a distributed and decentralized framework that uses the Ethereum blockchain for smart agriculture monitoring. IoT sensors collect crop-related data, which are analyzed by machine learning algorithms to offer predictions on crop demand, quality recommendations, and yield projections, and are securely stored on the blockchain [6].

Rahman et al. (2024) introduced an automated framework for paddy farming that combines IoT, blockchain technology, and Deep Learning (DL) algorithms. The system focuses on real-time pest identification using DL, integrates IoT devices for monitoring, and employs blockchain to secure data transmission, offering a robust solution for agricultural automation [15].

Osmanoglu et al. (2020) proposed a blockchain-based approach to yield estimation in agriculture, combining remote sensing techniques with traditional methods for accurate and robust crop yield estimation [7].

Patel et al. (2023) introduced an algorithm for recommending crops based on real-time soil attributes, leveraging soil sensors and blockchain for data validation and storage. Their user dashboard facilitates remote monitoring of farm practices and sensor statuses [8].

Ding et al. (2021) proposed an incentive mechanism for blockchain platforms to encourage the participation of IoT devices in mining, offering insights into strategic decision-making dynamics through simulations [9].

Kumar et al. (2023) introduced a deep learning architecture for an Intrusion Detection System (IDS) tailored for IoT-enabled Intelligent Agriculture (IA), integrating Contract-Assisted Secure Data Sharing (CASDS) with deep learning techniques for enhanced intrusion detection capabilities [10].

Yang et al. (2021) proposed a blockchain-based traceability system for agricultural product supply chains, ensuring secure data sharing and privacy protection through the implementation of Hyperledger Fabric [11].

In summary, these studies collectively contribute to the advancement of smart agriculture by leveraging statistical techniques, machine learning algorithms, and blockchain technology to address various challenges and improve efficiency, productivity, and sustainability in agricultural systems.

## 3   Proposed Methodology

Our proposed methodology amalgamates IoT sensors, machine learning (ML) algorithms, and Ethereum blockchain technology to improve the efficiency, precision, and security of agricultural data prediction. Strategically positioned IoT sensors, totaling seven in number, collect environmental data, which subsequently undergoes preprocessing and feature extraction before analysis by a pre-trained ML model. The integration of the Ethereum blockchain guarantees the integrity and security of data through encrypted transactions stored within an immutable and decentralized ledger. This holistic approach empowers stakeholders with actionable insights aimed at optimizing crop management strategies and resource allocation. Figure 1 delineates the schematic representation of the proposed methodology.

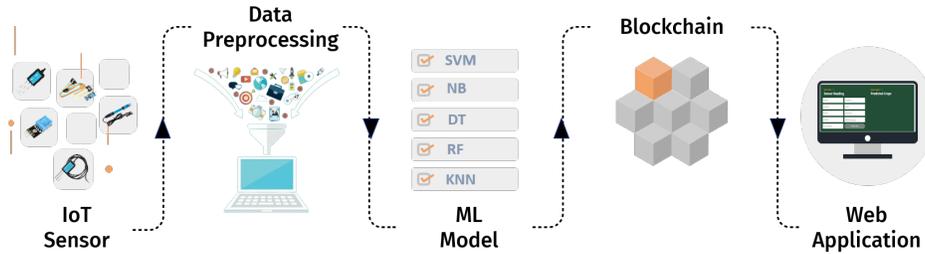

**Fig. 1.** Schematic Representation of the Proposed Methodology

The delineation of our system comprises five distinctive phases: Data Acquisition through IoT Sensors, Data Preprocessing, Deployment of the Machine Learning Model, Integration of Blockchain Technology, and Multi-Crop Prediction through a Web Application. Elucidation on each phase is furnished in subsequent subsections.

### 3.1 Data Acquisition

In our study, we used a comprehensive existing data set strategically designed to capture a wide range of environmental parameters crucial to crop growth and maturation. The dataset comprises 2200 instances, each meticulously annotated with seven essential features relevant to agricultural and environmental contexts. These features include Nitrogen (N), Phosphorus (P), Potassium (K), temperature, humidity, pH levels, and precipitation, all represented by the non-null float64 data type, ensuring there are no missing values. This rich dataset forms the foundation for our predictive modeling and analytical efforts, providing critical insights into the dynamics of crop cultivation and the impact of various environmental factors.

Although we use an existing dataset, this figure serves to contextualize how similar data might be collected through a network of strategically placed IoT sensors across an agricultural expanse. For instance:

– **Soil Nutrient Sensors:** Soil nutrient levels, including Nitrogen, Phosphorus, and Potassium, can be measured using ion-selective electrodes, which detect specific ions in the soil solution based on electrochemical potential. Additionally, spectroscopy-based sensors, such as near-infrared (NIR) sensors, analyze soil samples for nutrient concentration through spectral reflectance, offering rapid, noninvasive assessments.
– **Temperature and Humidity Sensors:** Digital sensors such as the DHT22, which incorporates a capacitive humidity sensor and a thermistor, provide precise readings of the environmental temperature and relative humidity. These sensors use ADC (Analog-to-Digital Conversion) to transmit data digitally, minimizing interference.

- **Soil pH Sensors:** Electrochemical soil pH sensors typically consist of a glass electrode sensitive to hydrogen-ion activity and a reference electrode. The voltage difference between the electrodes is translated into pH readings. These sensors are often paired with IoT modules for real-time monitoring.
- **Precipitation Sensors:** Rain gauges integrated with tipping-bucket mechanisms convert rainfall into electrical signals. IoT-enabled gauges often include additional components like flow meters to measure rain intensity and cumulative precipitation over time.

These sensors are interconnected using low-power wide-area network (LPWAN) technologies like LoRaWAN for efficient long-distance communication or Zigbee for localized sensor networks. The data collected is transmitted to a central cloud platform where it is processed using edge-computing frameworks or centralized machine learning models, enabling actionable insights into crop health and environmental conditions.

### 3.2 Data Preprocessing

In this section, our focus shifts towards data preprocessing techniques, with a particular emphasis on Min-Max normalization, tailored to render the data amenable for subsequent analysis and modeling endeavors.

Min-max normalization stands as a ubiquitous technique utilized for rescaling numerical features to a standardized range, typically spanning from 0 to 1. This normalization strategy proves particularly efficacious when confronted with features exhibiting disparate scales, thereby ensuring the equitable contribution of all features to the analytical framework.

### 3.3 Deployment of the Machine Learning Model

In this study, we evaluated several machine learning models for crop prediction, including Decision Trees (DT), Naive Bayes (NB), Support Vector Machines (SVM), k-Nearest Neighbors (KNN), Logistic Regression (LR), Random Forest (RF), and Neural Networks (NN). After rigorous testing, the Random Forest algorithm demonstrated the highest accuracy of 99.45

Random Forest, an ensemble learning method, works by constructing multiple decision trees during training and aggregating their predictions. This technique reduces overfitting and improves the model's generalization capability, particularly when dealing with complex, multi-dimensional datasets like agricultural sensor data.

In comparison, while Naive Bayes performed well with an accuracy of 99.36%, the Random Forest model consistently achieved better results across multiple evaluation metrics, including precision, recall, and F1-score. Given its robustness in handling both categorical and continuous features, Random Forest was selected as the optimal model for this system.

### 3.4 Blockchain: Ethereum

In our agricultural prediction system, the Ethereum blockchain plays a pivotal role in ensuring security, transparency, and data integrity by leveraging smart contracts. IoT sensors capture environmental data such as temperature, humidity, and soil moisture. This data is processed and securely recorded on the blockchain via smart contracts, ensuring its immutability and traceability. The blockchain provides a tamper-proof record of the data, preventing unauthorized modifications and ensuring that stakeholders can trust its accuracy.

The crop forecasts generated by the machine learning model are similarly recorded on the blockchain. This process not only secures the forecasts but also provides an auditable trail of when and how predictions were made. Stakeholder access to these predictions is logged as separate blockchain transactions, adding another layer of transparency.

Each transaction recorded on the blockchain includes essential components such as a nonce (to prevent replay attacks), gas price and limit (to manage transaction costs), recipient address, value (typically zero in this context), and encoded data specifying the smart contract function being executed (e.g., uploading sensor data or storing forecast results). These elements collectively ensure the secure and efficient functioning of the system.

For implementation, we use Solidity as the programming language to develop smart contracts. Tools such as Remix streamline contract development and testing, while MetaMask facilitates seamless browser-based interaction with the Ethereum blockchain. Users interact with the system through MetaMask, enabling them to execute transactions and view blockchain records in real time. Figure 2 illustrates the interaction between smart contracts and MetaMask. Furthermore, we utilize Ganache, a personal Ethereum blockchain, to simulate the network for development and testing purposes. Figure 3 presents Ganache's interface, depicting the creation of blockchain blocks and the recording of transactions.

This integration of the Ethereum blockchain ensures that the system achieves the highest standards of security, data integrity, and transparency while fostering trust among stakeholders.

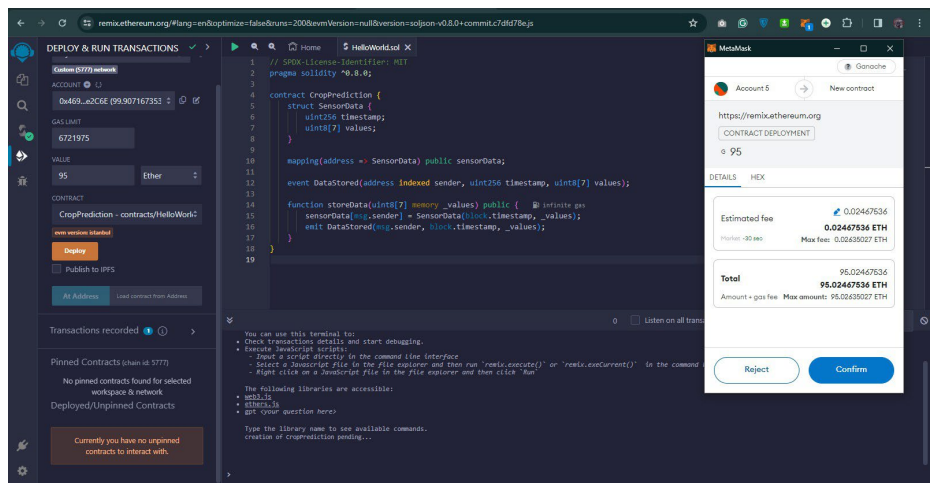

**Fig. 2.** Interaction between Smart Contact and MetaMask

**Fig. 3.** Blocks created on Ganache

The smart contract is pivotal to our crop prediction system, housing logic for receiving sensor data, executing prediction algorithms, and storing results on the blockchain. Algorithm 1 provides a high-level view of its design.

---

**Algorithm 1** Crop Prediction Smart Contract

---

**Contract** CropPrediction
**Attributes:** cropName, nitrogen, phosphorus, potassium, pH, rainfall, temperature, humidity
**Function:** addPrediction(cropName, n, p, k, ph, rain, temp, hum)
**Input:** cropName, n, p, k, ph, rain, temp, hum
**Action:** Create a new Prediction with the provided parameters and add it to the predictions array
**Function** getPrediction(uint256 index) → (string, uint256, uint256, uint256, uint256, uint256, uint256, uint256)
**Input:** index: uint256
**Output:** cropName: string, n: uint256, p: uint256, k: uint256, ph: uint256, rain: uint256, temp: uint256, hum: uint256
**Action:** Retrieve the Prediction at the specified index from the predictions array and return its attributes

---

### 3.5 Multiple Crop Prediction and Web Application

To facilitate crop prediction from sensor data, we developed a user-friendly web interface using the Django framework. This interface enables stakeholders to access crop projections and recommendations generated by a pre-trained machine-learning model stored as a .pkl file. Key features of the interface include:

1. **Real-Time Data Display:** Visualizations of environmental data (e.g., temperature, humidity) captured from IoT sensors.
2. **Crop Recommendations:** A ranked list of the top crop predictions based on environmental conditions and historical data.
3. **Blockchain Integration:** Seamless interaction with Ethereum blockchain via Meta-Mask, allowing stakeholders to view tamper-proof records of sensor data and predictions.

## 4    Results and Analysis

This section presents the findings from our experimentation with machine learning models for agricultural decision support. We assess the performance of various algorithms, both tuned and untuned, in predicting suitable crops based on sensor data. Additionally, we introduce a user-friendly web interface designed to provide farmers with actionable insights. Finally, we offer a comparative analysis between our proposed system and existing approaches in the field.

### 4.1   Model Performance Comparison

In evaluating the performance of machine learning (ML) models, experiments were conducted both without tuning and with tuning processes over the existing dataset. This involved the exploration of different parameters of various models to optimize results. The evaluation metrics considered include accuracy, precision, recall, and F1 score.

Fig 4 illustrates the accuracy comparison among all ML models. Each model is represented by an abbreviation: "DT" for Decision Tree, "NB" for Gaussian Naive Bayes, "SVM" for Support Vector Machine, "LR" for Logistic Regression, "RF" for Random Forest, "KNN" for K-Nearest Neighbors, and "NN" for Neural Network.

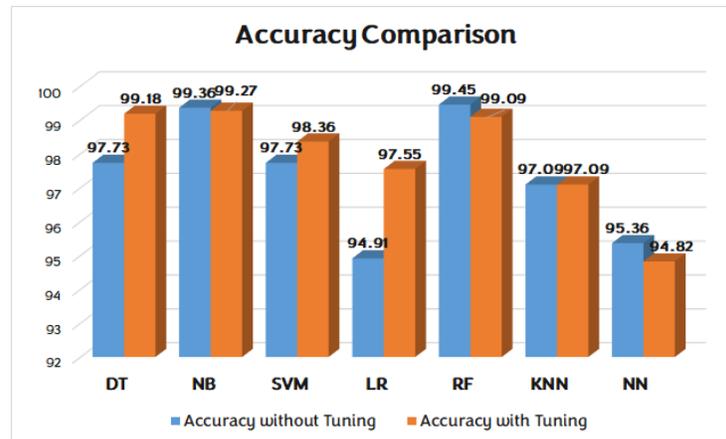

**Fig. 4.** Accuracy comparison among all ML models

Table 1 presents the comprehensive findings of all ML models across different parameters, including training time, testing time, accuracy, precision, and recall.

**Table 1.** Model Analysis without Tuning

| Algorithm | Training Time | Testing Time | Accuracy (%) | Precision (%) | Recall (%) |
|---|---|---|---|---|---|
| Decision Tree | 0.179 | 0.005 | 97.73 | 97.71 | 97.72 |
| Gaussian Naive Bayes | 0.012 | 0.007 | 99.36 | 99.34 | 99.29 |
| Support Vector Machine | 0.042 | 0.029 | 97.73 | 97.74 | 97.61 |
| Logistic Regression | 0.153 | 0.002 | 94.91 | 94.79 | 94.77 |
| **Random Forest** | **0.213** | **0.014** | **99.45** | **99.45** | **99.38** |
| K-Nearest Neighbors | 0.008 | 0.073 | 97.09 | 97.22 | 96.97 |
| Neural Network | 1.706 | 0.005 | 95.36 | 95.47 | 95.15 |

### 4.2 Web Interface for Agricultural Applications

Within agricultural contexts, a sophisticated Django-powered web interface has been crafted to streamline user interaction for inputting sensor data encompassing soil moisture, temperature, and humidity. Employing advanced algorithms, this platform meticulously scrutinizes the provided data to forecast optimal crops for cultivation, thereby advocating for sustainable agricultural practices and endeavoring to optimize yields (see Figure 5).

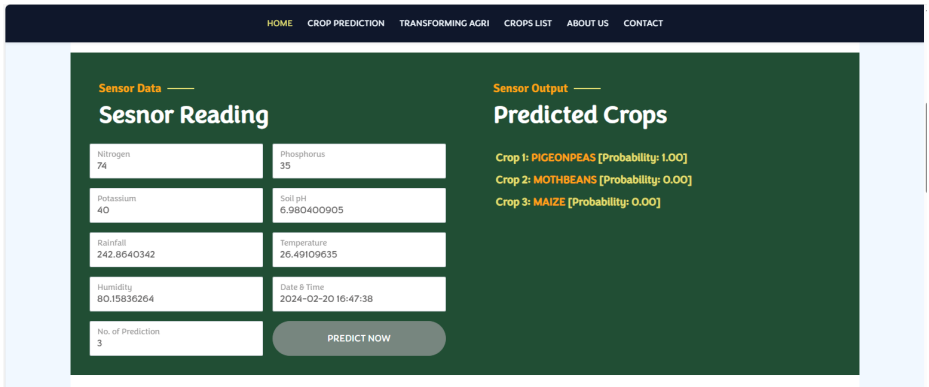

**Fig. 5.** Web Interface with Output

### 4.3 Comparative Analysis

A comparative analysis of existing systems with the proposed system is provided in Table 2. This table offers an overview of various models utilized, their corresponding accuracies, whether feature selection or engineering techniques were employed, and the crops predicted by each model.

Table 2. Comparative analysis of some existing systems with proposed system

| Ref. | Model Used | Accuracy | BC Used | Predicted Crops |
|---|---|---|---|---|
| [2] | DT | 97.28% | No | Single |
|  | KNN | 97.05% |  |  |
|  | RF | 98.33% |  |  |
|  | NB | 99.12% |  |  |
|  | SVM | 93.71% |  |  |
| [3] | Decision Tree | 81% | No | Single |
|  | KNN | 88% |  |  |
|  | LR | 88.26% |  |  |
|  | NB | 82% |  |  |
|  | NN | 89.88% |  |  |
|  | SVM | 78% |  |  |
| [12] | KNN | 97.04% | No | Single |
|  | DT | 97.95% |  |  |
|  | RF | 99.32% |  |  |
| [13] | NB | 99.46% | No | Single |
|  | LR | 97.99% |  |  |
|  | Simple Logistic | 98.66% |  |  |
|  | Regression | 98.38% |  |  |
|  | DT | 88.50% |  |  |
|  | Random Tree | 98.12% |  |  |
| [14] | KNN | 75.55% | No | Single |
|  | NB | 91.11% |  |  |
| Proposed System | DT | 97.73% | Yes | Multiple |
|  | NB | 99.36% |  |  |
|  | SVM | 97.73% |  |  |
|  | LR | 94.91% |  |  |
|  | RF | 99.45% |  |  |
|  | KNN | 97.09% |  |  |
|  | NN | 95.36% |  |  |

## 5 Conclusion

This study demonstrated the potential of integrating IoT, machine learning, and blockchain technology to improve crop forecasting in agriculture. The Random Forest model, in particular, achieved the highest accuracy, providing robust predictions based on real-time environmental data collected from sensors. The use of the Ethereum blockchain ensured the security and integrity of sensor data, offering an immutable, tamper-proof record for stakeholders.

Adopting this integrated approach to crop forecasting can yield significant economic benefits to farmers. By providing accurate and timely crop predictions, the sys-tem enables farmers to make informed decisions about crop selection, planting schedules, and resource allocation. This can reduce input costs such as water, fertilizers, and pesticides by optimizing their use based on precise forecasts. In addition, blockchain integration fosters transparency and trust in supply chains, potentially opening new market opportunities for farmers and increasing their bargaining power. These advantages collectively contribute to improved yields, higher profitability, and reduced financial risks.

Future work will focus on optimizing and expanding the system to address additional challenges in agriculture. Key areas of future research and enhancements include the following:

- **Integration of Real-Time Sensor Networks:** Enabling live data feeds for dynamic forecasting that adapts to changing environmental conditions.
- **Incorporation of Additional Data Sources:** Including external factors such as market trends, pest populations, pesticide levels, and disease outbreaks to enhance predictive accuracy and relevance.
- **Exploration of Alternative Blockchain Platforms:** Investigating other blockchain technologies, such as Hyperledger or Polygon, to assess their suitability for agricultural applications and potentially improve scalability and cost-efficiency.
- **AI-Driven Recommendations and User Feedback:** Developing advanced AI algorithms to provide actionable insights tailored to farmers' specific needs, and incorporating feedback mechanisms to refine predictions and recommendations over time.
- **Mobile Accessibility:** Creating a mobile-friendly interface to ensure widespread usability and access for farmers in remote areas.

This integrated approach will contribute to a more sustainable and efficient agricultural sector while improving the livelihoods of farmers.

**Acknowledgments.** The research was supported by the ITRRC (Information Technology Research and Resource Center) Cybersecurity Research Lab, Jagannath University, Dhaka, Bangladesh (https://itrrc.com/).